\documentclass[runningheads]{llncs}

\usepackage[T1]{fontenc}
\usepackage{graphicx}
\usepackage{microtype}
\usepackage{booktabs} 

\usepackage{amssymb}
\usepackage{mathtools}
\usepackage{amsmath}
\usepackage{subfig}
\usepackage{paralist}
\usepackage{tabularx}
\usepackage{marvosym}
\usepackage{xcolor}
\usepackage{ulem}
\usepackage{svg}
\usepackage[capitalize,noabbrev]{cleveref}
\usepackage[textsize=tiny]{todonotes}

\def\our{CoHiClust}
\def\loss{\mathrm{CoHiLoss}}

\def\R{\mathbb{R}}

\begin{document}
\title{Contrastive Hierarchical Clustering}
%
%


\author{Micha\l{} Znalezniak\inst{1} \and
Przemys\l{}aw Rola\inst{2,4} \and
Patryk Kaszuba\inst{3} \and 
Jacek Tabor\inst{4} \and
Marek Śmieja\inst{4}(\Letter)}
%
%
\institute{Advanced Micro Devices, Inc.\\
\and
Institute of Quantitative Methods in Social Sciences, Cracow University of Economics\\
\and
Faculty of Mathematics and Computer Science, Adam Mickiewicz University\\
\and 
Faculty of Mathematics and Computer Science, Jagiellonian University, \\Krak\'ow, Poland
\\
\email{mznalezn@amd.com}, 
\email{marek.smieja@uj.edu.pl},
}
\maketitle              
\begin{abstract}
Deep clustering has been dominated by flat models, which split a dataset into a predefined number of groups. Although recent methods achieve an extremely high similarity with the ground truth on popular benchmarks, the information contained in the flat partition is limited. In this paper, we introduce \our{}, a Contrastive Hierarchical Clustering model based on deep neural networks, which can be applied to typical image data. By employing a self-supervised learning approach, \our{} distills the base network into a binary tree without access to any labeled data. The hierarchical clustering structure can be used to analyze the relationship between clusters, as well as to measure the similarity between data points. Experiments demonstrate that \our{} generates a reasonable structure of clusters, which is consistent with our intuition and image semantics. Moreover, it obtains superior clustering accuracy on most of the image datasets compared to the state-of-the-art flat clustering models. Our implementation is available at \url{https://github.com/MichalZnalezniak/Contrastive-Hierarchical-Clustering}.

\end{abstract}

\section{Introduction}

Clustering, a fundamental branch of unsupervised learning, is often one of the first steps in data analysis, which finds applications in anomaly detection \cite{barai2017outlier}, personalized recommendations \cite{zhang2014taxonomy} or bioinformatics \cite{lakhani2015clustering}. Since it does not use any information about class labels, representation learning becomes an integral part of deep clustering methods. Initial approaches use representations taken from pre-trained models \cite{guerin2017cnn,naumov2021objective} or employ autoencoders in joint training of the representation and the clustering model \cite{guo2017improved,mautz2019deep}. Recent models designed to image data frequently follow the self-supervised learning principle, where the representation is trained on pairs of similar images generated automatically by data augmentations \cite{li2021contrastive,dang2021nearest}.
Since augmentations used for image data are class-invariant, the latter techniques of ten obtain a very high similarity to the ground truth classes. However, we should be careful when comparing clustering techniques only by inspecting their accuracy with ground truth classes because the primary goal of clustering is to deliver information about data and not to perform classification.

\begin{figure*}[!htb]
    \centering
    \includegraphics[width=\textwidth]{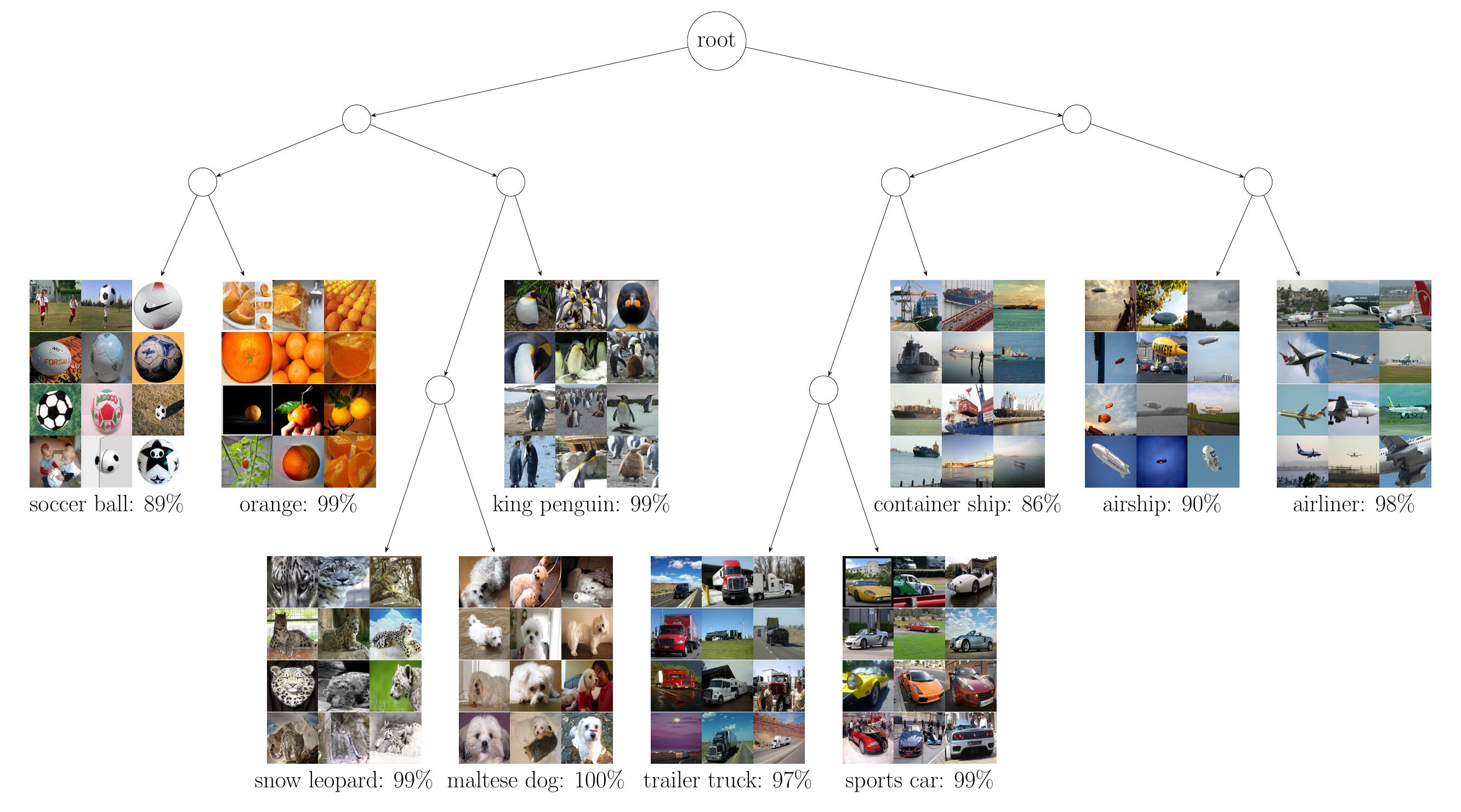}
    \caption{{\bf A hierarchy generated by \our{} for ImageNet-10.} As can be seen, \our{} reliably reflected the ground-truth classes in leaf nodes. In addition to information delivered by flat partition, in hierarchical models neighbor leaves contain images sharing similar characteristic. It is evident that images with soccer ball are similar to pictures with oranges because of their shapes. Dogs are more similar to leopards than to penguins, which is reflected in the constructed hierarchy. The same hold when analyzing the leafs representing cars, trucks and ships. Looking at the first hierarchy level, we observe a distinction on the right sub-tree representing machines and left-sub-tree dominated by animals. Moreover, balls and oranges are separated from the animal branch.}
    \label{fig:imagenet}
\end{figure*}

Most of the works in the area of deep clustering focus on producing flat partitions with a predefined number of groups. Although hierarchical clustering has gained considerable attention in classical machine learning and has been frequently applied in real-life problems \cite{zou2020sequence,smieja2014asymmetric}, its role has been drastically marginalized in the era of deep learning. In the case of hierarchical clustering, the exact number of clusters does not have to be specified because we can inspect the partition at various tree levels. Moreover, we can analyze the clusters' relationships, e.g. by finding superclusters or measuring the distance between groups in the hierarchy. These advantages make hierarchical clustering an excellent tool for analyzing complex data. However, to take full advantage of hierarchical clustering, it is necessary to create an appropriate image representation, which is possible due to the use of deep neural networks. To our knowledge, DeepECT \cite{mautz2019deep,mautz2020deepect} is the only hierarchical clustering model trained jointly with the neural network. 
However, this method has not been examined for larger datasets of color images.

To fill this gap, we introduce \our{} ({\bf Co}ntrastive {\bf Hi}erarchical {\bf Clust}ering), which creates a cluster hierarchy and works well on typical color image databases. \our{} uses a neural network to generate a high-level representation of data, which is then distilled into the tree hierarchy by applying the projection head, see Figure \ref{fig:diagram}. The whole framework is trained jointly in an end-to-end manner without labels using our novel contrastive loss and data augmentations generated automatically following the self-supervised learning principle.

The constructed hierarchy uses the structure of a binary tree, where the sequence of decisions made by the internal nodes determines the final assignment to the clusters (leaf nodes). In consequence, similar examples are processed longer by the same path than dissimilar ones. By inspecting the number of edges needed to connect two clusters (leaves), we obtain a natural similarity measure between data points. By applying a pruning strategy, which removes the least informative leaf nodes, we can restrict the hierarchy to a given number of leaves and fine-tune the whole hierarchy.



The proposed model has been examined on various image datasets and compared with hierarchical and flat clustering baselines. By analyzing the constructed hierarchies, we show that \our{} generates a structure of clusters that is consistent with our intuition and image semantics, see Figures \ref{fig:imagenet} and \ref{fig:cifar} for illustration and discussion. Our analysis is supported by a quantitative study, which shows that \our{} outperforms the current clustering models on 5 out of 7 image datasets, see Tables \ref{tab:large} and \ref{tab:hier}.


Our main contributions are summarized as follows:
\begin{compactitem}
    \item We introduce a hierarchical clustering model \our{}, which converts the base neural network into a binary tree. The model is trained effectively without supervision using our novel hierarchical contrastive loss applied to self-generated data augmentations.
    \item We conducted an extensive experimental study, which confirms that \our{} is very competitive with current state-of-the-art flat clustering models. Moreover, we show that it builds hierarchies based on well-defined and intuitive patterns retrieved from the data.
    \item Since \our{} is the first deep hierarchical clustering model applied to color image datasets, we provide a benchmark, which can be used to compare hierarchical clustering methods.
\end{compactitem}

\section{Related Work}

In this section, we briefly introduce some recent developments on three related topics: contrastive learning, deep clustering, and hierarchical methods.

\paragraph{Contrastive Learning} The basic idea of contrastive learning is to learn a feature space in which similar pairs stay close to each other, while dissimilar ones are far apart \cite{chopra2005learning}. 
In recent works, it was observed that in selected domains, such as computer vision, positive (similar) pairs can be generated automatically using adversarial perturbations \cite{miyato2018virtual} or data augmentation \cite{Moco_2020}, giving rise to a new field called self-supervised learning \cite{simclr}. Fine-tuning a simple classifier 
on self-supervised representation allows for obtaining the accuracy comparable to a fully supervised setting.
SimCLR \cite{Moco_2020} applies NT-Xent loss to maximize the agreement between differently augmented views of the same sample. Barlow Twins \cite{zbontar2021barlow} learns to make the cross-correlation matrix between two distorted versions of the same samples close to the identity. BYOL \cite{grill2020bootstrap} claims to achieve new state-of-the-art results without using negative samples. Other works use memory banks to reduce the cost of computing the embeddings of negative samples in every batch \cite{wu2018unsupervised,Moco_2020}.

\paragraph{Deep clustering} A primary focus in deep embedded clustering has merely been on flat clustering objectives with the actual number of clusters known a priori. DEC \cite{xie2016unsupervised} is one of the first works, which combines the auto-encoder loss with a clustering objective to jointly learn the representation and perform clustering. This idea was further explored with some improvements in IDEC \cite{guo2017improved}, JULE \cite{yang2016joint} and DCEC \cite{guo2017deep}. IMSAT \cite{hu2017learning} and IIC \cite{ji2019invariant} use perturbations to generate pairs of similar examples and apply information-theoretic objectives for training. PICA \cite{huang2020deep} maximizes the global partition confidence of the clustering solution to find the most semantically plausible data separation. Following progress in self-supervised learning, CC \cite{li2021contrastive} and DCSC \cite{Zhang2022} perform contrastive learning by generating pairs of positive and negative instances through data augmentations. 




\paragraph{Hierarchical methods} 

Hierarchical clustering algorithms are a well-established area within classical data mining \cite{Murtagh}, but have rarely been studied in deep learning. DeepECT \cite{mautz2019deep,mautz2020deepect} is the only method that jointly learns deep representation using autoencoder architecture and performs hierarchical clustering in a top-down manner. Unfortunately, no comparative study has been conducted on large image data. The experimental study of objective-based hierarchical clustering methods performed on the embedding vectors of pre-trained deep learning models is presented in \cite{naumov2021objective}. In the case of classification, there is a growing interest in deep hierarchical methods, which in our opinion should also be reflected in the area of unsupervised learning. SDT \cite{frosst2017distilling} is one of the first models that distills the base neural networks into a soft decision tree. More advanced methods automatically generate deep networks with a tree structure in a multistep or end-to-end manner \cite{tanno2019adaptive,alaniz2021learning,wan2020nbdt}.







\section{\our{} model}

The proposed \our{} builds a hierarchy of clusters based on the output of the base neural network. 
There are three key components of \our{}:
\begin{compactitem}
    \item The base neural network that generates the representation used by the tree.
    \item The tree model, which assigns data points to clusters by a sequence of decisions.
    \item The regularized contrastive loss, which allows for training the whole framework. 
\end{compactitem}
We discuss the above components in detail in the following parts.

\subsection{Tree hierarchy}

We use a soft binary decision tree to create a hierarchical structure, where leaves play the role of clusters (similar to \cite{frosst2017distilling}). In contrast to hard decision trees, every internal node defines the probability of taking a left/right branch. The final assignment of the input examples to clusters involves partial decisions made by the internal nodes. Aggregating these decisions induces the posterior probability over leaves. 

\begin{figure*}[!htb]
    \centering
    \includegraphics[width=\textwidth]{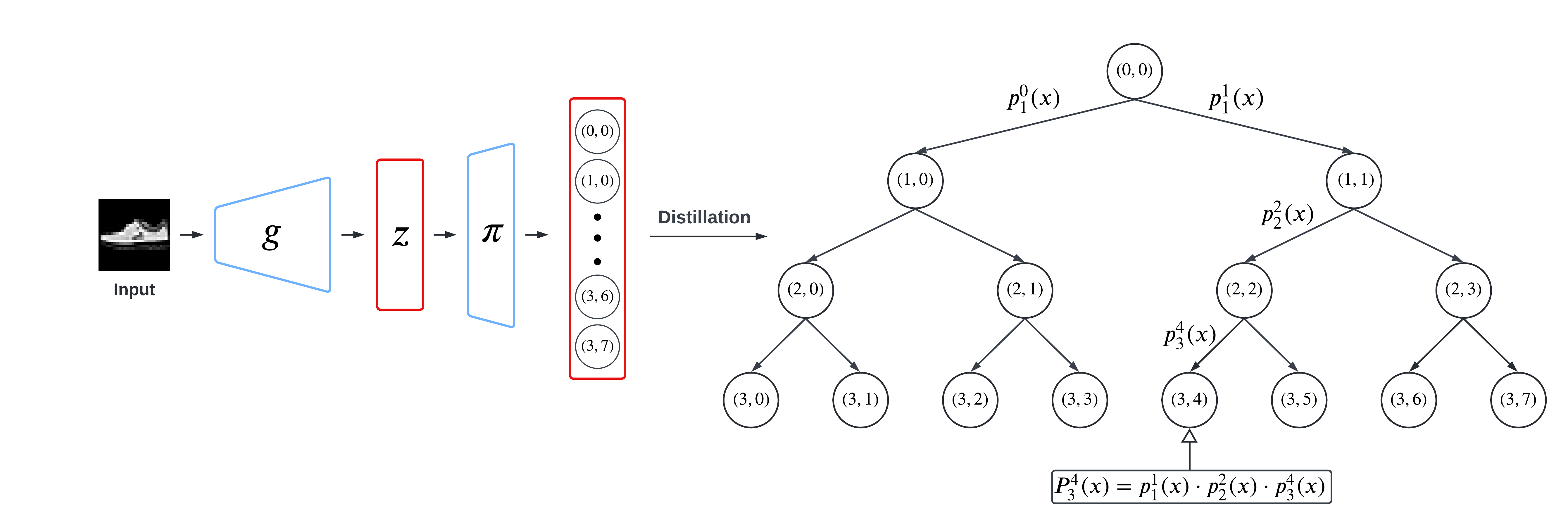}
    \caption{{\bf Illustration of \our{}.} The output neurons of the projection head $\pi$ (appended to the base network $g$) model decisions made by the internal tree nodes. The final assignment of the input example to the cluster (leaf node) is performed by aggregating edge probabilities located on the path from the root to this leaf. 
    }
    \label{fig:diagram}
\end{figure*}

Let us consider a complete binary tree with $T$ levels, where the root is located at the level $0$ and the leaves are represented at the level $T$. This gives us $2^t$ nodes at the level $t$ denoted by tuples $(t,i)$, for $i=0,1,\ldots, 2^t-1$, see Figure~\ref{fig:diagram}. The path from root to node $(t,i)$ is given by the sequence of binary decisions $y=(y_1,\ldots,y_t) \in \{0,1\}^t$ made by the internal nodes, where $y_s=0$ ($y_s=1$) means that we take the left (right) branch that is at the node at the level $s$. Observe that we can retrieve the index $j$ of the node at the level $s$ from $y$ by taking ${j=b_s(y) = \sum_{m=1}^s y_m 2^{s-m}}$. In other words, the first $s$ bits of $y$ are a binary representation of the number $j$.

We consider the path induced by the sequence of decisions $y=(y_1,\ldots,y_{t}) \in \{0,1\}^t$, which goes from the root to the node $(t,i)$, where $i=b_{t}(y)$. We want to calculate the probability $P_t^i(x)$ that the input example $x \in \R^D$ reaches node $(t,i)$. If $p_{s}^{b_s(y)}(x)$ is the probability of going from the parent node $(s-1,b_{s-1}(y))$ to its descendant $(s,b_{s}(y))$, then
$$
P_t^i(x) = p_1^{b_1(y)}(x) \cdot p_2^{b_2(y)}(x) \cdot \ldots p_t^{b_t(y)}(x).
$$
Observe that $P_t(x) = [P_t^0(x), P_t^1(x), \ldots, P_t^{2^t-1}(x)]$ defines a proper probability distribution, i.e. ${\sum_{j=0}^{2^t-1} P_t^j(x) = 1}$. As a consequence, the probability distribution on the clusters (leaves) is equal to $P_T(x)$, see Figure~\ref{fig:diagram}.

\subsection{Tree generation}

To generate our tree model, we need to parameterize the probabilities $p_t^i(x)$ of taking the left/right branch in every internal node. To this end, we employ a neural network $g: \R^D \to \R^N$ with an additional projection head $\pi: \R^N \to \R^K$, where $K = 2^{T}-1$ and $T$ is the height of the tree. The number $K$ of output neurons equals the number of internal tree nodes.  

The neural network $g$ is responsible for extracting high-level information from the data. It can be instantiated by a typical architecture, such as ResNet, and is used to generate embeddings $z=g(x)$ of the input data. 

The projection head $\pi$ operates on the embeddings $z$ and parameterizes the decisions made by the internal tree nodes. In our case, $\pi$ is a single layer network with the output dimension equal to the number of internal nodes. To model binary decisions of internal nodes, we apply the sigmoid function $\sigma$. Consequently, the projection head is given by $\pi(z) = [\sigma (w_1^T z + b_1), \ldots, \sigma (w_K^T z + b_K)]$, where $w_n \in \R^N$ and $b_n \in \R$ are trainable parameters of $\pi$. By interpreting the output neurons of $\pi$ as the internal nodes of the decision tree, we obtain the probabilities of the left edges in the nodes:
$$
p_{t+1}^{2i}(x) = \sigma(w_n^T z + b_n) \text{ , for } n = 2^t+i.
$$
Note that $p_{t+1}^{2i-1}(x) = 1 - p_{t+1}^{2i}(x)$ always corresponds to the probability of the right edge.


\subsection{Contrastive hierarchical loss}

To train \our{}, we introduce the hierarchical contrastive loss function designed for trees. Our idea is based on maximizing the likelihood that similar data points will follow the same path. The more similar data points, the longer they should be routed through the same nodes. Since we work in an unsupervised setting, we use a self-supervised approach and generate similar images using data augmentations.

Let us consider two data points $x_1, x_2$ and their posterior probabilities $P_t(x_1), P_t(x_2)$ at the level $t$. The probability that $x_1$ and $x_2$ reach the same node at this level is given by the scalar product $P_t(x_1) \cdot P_t(x_2)= \sum_{i=0}^{2^t-1} P_t^i(x_1) P_t^i(x_2)$. This term is maximal if both probabilities are identical one-hot vectors. 
In a training phase, we do not want to force hard splits in the nodes (binary probabilities), because in this way the model quickly finds a local minimum by assigning data points to fixed leaves with high confidence. Instead of sticking to hard assignments in a few training epochs, we want to let the model explore possible solutions. To this end, we take the square root before applying the scalar product, which corresponds to the Bhattacharyya coefficient \cite{bhattacharyya1946measure}:
\begin{equation} \label{eq:square}
s_t(x_1,x_2) = \sqrt{P_t(x_1) \cdot P_t(x_2)}= \sum_{i=0}^{2^t-1} \sqrt{P_t^i(x_1) P_t^i(x_2)}.
\end{equation}
Observe that $s_t(x_1,x_2) = 1$, if only $P_t(x_1) = P_t(x_2)$ (probabilities do not have to binarize), which leads to the exploration of possible paths. By aggregating the similarity scores over all tree levels, we arrive at our final similarity function $s(x_1,x_2) = \sum_{t=0}^{T-1} s_t(x_1,x_2)$.



In a training phase, we take a minibatch $\{x_j\}_{j=1}^N$ of $N$ examples and generate its augmented view $\{\tilde{x}_j\}_{j=1}^N$. Every pair $(x_j,\tilde{x}_j)$ is considered positive, which means that we will maximize their similarity score. As a consequence, we encourage the model to assign them to the same leaf node. To avoid degenerate solutions, where all points end up in the same leaf, we treat all other pairs as negative and minimize their similarity scores. Finally, the proposed hierarchical contrastive loss is given by:
$$
\loss = \frac{1}{N(N-1)} \sum_{j=1}^N \sum_{i\neq j} s(x_j,\tilde{x}_{i}) - \frac{1}{N} \sum_{j=1}^N s(x_j,\tilde{x}_j).
$$
By minimizing the above loss, we maximize the likelihood that similar data points follow the same path (second term) and minimize the likelihood that dissimilar ones are grouped together.

\subsection{Regularization}


The final cluster assignments are induced by aggregating several binary decisions made by the internal tree nodes. In practice, the base neural network may not train all nodes and, in consequence, use only a few leaves for clustering. While selecting the number of clusters in flat clustering is desirable, here we would like to create a hierarchy, which is not restricted to a given number of leaves. To enable end-to-end training of the base neural network with an arbitrary number of leaves, we consider two regularization strategies. 

The first regularization (dubbed $R_1$) explicitly encourages the model to use both left and right sub-trees equally \cite{frosst2017distilling}. We realize this postulate by minimizing the cross-entropy between the desired distribution $[0.5, 0.5]$ and the actual distribution to choose the left or right path in a given node. 

The second regularization (dubbed $R_2$) does not directly influence the routing in the tree, but focuses on improving the output representation of the base network $g$. For this purpose, we use a projection head $\phi:\R^N \to \R^M$, which transforms the embeddings $z=g(x)$ of the input data, and apply the NT-Xent loss \cite{simclr} to $\tilde{z} = \phi(z)$. With the NT-Xent loss, we maximize the cosine similarity for all positive pairs and minimize the cosine similarity for all negative pairs. For simple datasets, such as MNIST or F-MNIST, $\phi$ is an identity function, while for more complex color images, it is a two-layer network.

Taking together the contrastive loss $\loss$ with two regularization functions $R_1$ (for entropy) and $R_2$ (for NT-Xent), we arrive at our final loss:
\begin{align} \label{eq:loss}
\mathrm{Loss} = & \loss + \beta_1 R_1 + \beta_2 R_2,
\end{align}
where $\beta_1, \beta_2$ are the hyperparameters that define the importance of the regularization terms $R_1$ and $R_2$, respectively. To generate a complete hierarchy (complete tree with assumed height), we set $\beta_1$ proportional to the depth of the tree $\beta_1 = 2^{-T}$ \cite{frosst2017distilling} and $\beta_2 = 1$ \cite{Li2021contrastive_clustering}.


\subsection{Training}

\our{} can be trained end-to-end by minimizing \eqref{eq:loss}. We verified that training is even more effective when we introduce a pre-training phase of the base neural network. To this end, we perform self-supervised representation learning of the base network $g$ using $R_2$ regularization (it corresponds to the SimCLR model \cite{simclr}). In \Cref{sec:analysis}, we show that pre-training allows us to reduce the number of training epochs and leads to a better overall performance of \our{}. 

The proposed model builds a complete tree with $2^T$ leaves. Although such a structure is useful for analyzing the hierarchy of clusters, in some cases we are interested in creating a tree with the requested number of groups. For this purpose, we apply a pruning step that reduces the least significant leaf nodes. Namely, we reduce leaves with the lowest expected fraction of data points: $P_T^i = \frac{1}{|X|} \sum_{x \in X} P_T^i(x)$. Pruning is realized after a few first training epochs of \our{} (after the pre-training phase). We remove one leave per epoch. Next, \our{} is trained with the final number of leaves.




\section{Experiments}

\begin{table}[t]
\centering
\caption{Comparison with flat clustering methods on datasets of color images.
} \label{tab:large}
\tiny
\begin{tabular}{lccccccccccccccc}
\toprule
Dataset & \multicolumn{3}{c}{CIFAR-10} & \multicolumn{3}{c}{CIFAR-100} & \multicolumn{3}{c}{STL-10} & \multicolumn{3}{c}{ImageNet-10} & \multicolumn{3}{c}{ImageNet-Dogs} \\
\midrule
Metrics & NMI & ACC & ARI & NMI & ACC & ARI & NMI & ACC & ARI & NMI & ACC & ARI & NMI & ACC & ARI \\
\midrule
K-means \cite{MacQueen} & 0.087 & 0.229 & 0.049 & 0.084 & 0.130 & 0.028 & 0.125 & 0.192 & 0.061 & 0.119 & 0.241 & 0.057 & 0.055 & 0.105 & 0.020 \\
SC \cite{Zelnik2004} & 0.103 & 0.247 & 0.085 & 0.090 & 0.136 & 0.022 & 0.098 & 0.159 & 0.048 & 0.151 & 0.274 & 0.076 & 0.038 & 0.111 & 0.013 \\
AC \cite{Gowda1978AgglomerativeCU} & 0.105 & 0.228 & 0.065 & 0.098 & 0.138 &  0.034  & 0.239 & 0.332 & 0.140 & 0.138 & 0.242 & 0.067 & 0.037 & 0.139 & 0.021 \\
NMF \cite{Cai2009} & 0.081 & 0.190 & 0.034 & 0.079 & 0.118 & 0.026  & 0.096 & 0.180 & 0.046 & 0.132 & 0.230 & 0.065 & 0.044 & 0.118 & 0.016 \\
AE \cite{Bengio2006} & 0.239 & 0.314 & 0.169 & 0.100 & 0.165 & 0.048  & 0.250 & 0.303 & 0.161 & 0.210 & 0.317 & 0.152 & 0.104 & 0.185 & 0.073 \\
DAE \cite{vincent2010} & 0.251 & 0.297 & 0.163 & 0.111 & 0.151 & 0.046 & 0.224 & 0.302 & 0.152 & 0.206 & 0.304 & 0.138 & 0.104 & 0.190 & 0.078 \\
DCGAN \cite{Radford2015} & 0.265 & 0.315 & 0.176 &  0.120 & 0.151 & 0.045 & 0.210 & 0.298 & 0.139 & 0.225 & 0.346 & 0.157 & 0.121 & 0.174 & 0.078 \\
DeCNN \cite{Zeiler2010} & 0.240 & 0.282 & 0.174 & 0.092 & 0.133 & 0.038 & 0.227 & 0.299 & 0.162 & 0.186 & 0.313 & 0.142 & 0.098 & 0.175 & 0.073 \\
VAE \cite{kingma2013} & 0.245 & 0.291 & 0.167 & 0.108 & 0.152 & 0.040 & 0.200 & 0.282 & 0.146 & 0.193 & 0.334 & 0.168 & 0.107 & 0.179 & 0.079 \\
JULE \cite{yang2016joint} & 0.192 & 0.272 & 0.138 &  0.103 & 0.137 & 0.033 & 0.182 & 0.277 & 0.164 & 0.175 & 0.300 & 0.138 & 0.054 & 0.138 & 0.028 \\
DEC \cite{xie2016unsupervised} & 0.257 & 0.301 & 0.161 &  0.136 & 0.185 & 0.050 & 0.276 & 0.359 & 0.186 & 0.282 & 0.381 & 0.203 & 0.122 & 0.195 & 0.079 \\
DAC \cite{Chang_2017_ICCV} & 0.396 & 0.522 & 0.306 & 0.185 & 0.238 & 0.088 &  0.366 & 0.470 & 0.257 & 0.394 & 0.527 & 0.302 & 0.219 & 0.275 & 0.111 \\
DCCM \cite{Wu_2019_ICCV} & 0.496 & 0.623 & 0.408 & 0.285 & 0.327 & 0.173 & 0.376 & 0.482 & 0.262 & 0.608 & 0.710 & 0.555 & 0.321 & 0.383 & 0.182 \\
PICA \cite{huang2020deep} & 0.591 & 0.696 & 0.512 & 0.310 & 0.337  & 0.171  & 0.611 & 0.713 & 0.531 & 0.802 & 0.870 & 0.761 & 0.352 & 0.352 & 0.201 \\
CC \cite{Li2021contrastive_clustering} & 0.705 & 0.790 & 0.637 & 0.431 & 0.429 & 0.266 & \textbf{0.764} & \textbf{0.850} & \textbf{0.726} & 0.859 & 0.893 & 0.822 & \textbf{0.445} & \textbf{0.429} & \textbf{0.274}\\
{\bf \our{}} & \textbf{0.779} & \textbf{0.839} & \textbf{0.731} & \textbf{0.467} & \textbf{0.437} & \textbf{0.299}  & 0.584 & 0.613 & 0.474 & \textbf{0.907} & \textbf{0.953} & \textbf{0.899} & 0.411 & 0.355 & 0.232 \\ 
\bottomrule
\end{tabular}
\end{table}

First, we evaluate our method on several datasets of color images of various resolutions and with a diverse number of classes. In addition to reporting similarity scores with ground-truth partitions, we analyze the constructed hierarchies, which in our opinion is equally important in practical use-cases. Next, we perform an ablation study and investigate the properties of \our{}. Finally, we compare \our{} with existing hierarchical methods. Details of the experimental settings and additional results are included in the Appendix.


\subsection{Clustering color images}

\paragraph{Benchmark} We perform the evaluation on typical bemchmark datasets: CIFAR-10, CIFAR-100, STL-10, ImageNet-Dogs, and ImageNet-10, see the Appendix for their summary. We process each data set at its original resolution. Since none of the previous hierarchical methods have been examined on these datasets, we use the benchmark that includes the flat clustering methods reported in \cite{Li2021contrastive_clustering}. According to previous works in contrastive clustering, \our{} uses ResNet architectures as a backbone network. To measure the similarity of the constructed partition with the ground truth, we apply three widely-used clustering metrics: normalized mutual information (NMI), clustering accuracy (ACC), and adjusted rand index (ARI). In the Appendix, we also show the DP (dendrogram purity) of the hierarchies generated by \our{}.

\begin{figure}[!htb]
    \centering
    \includegraphics[width=\textwidth]{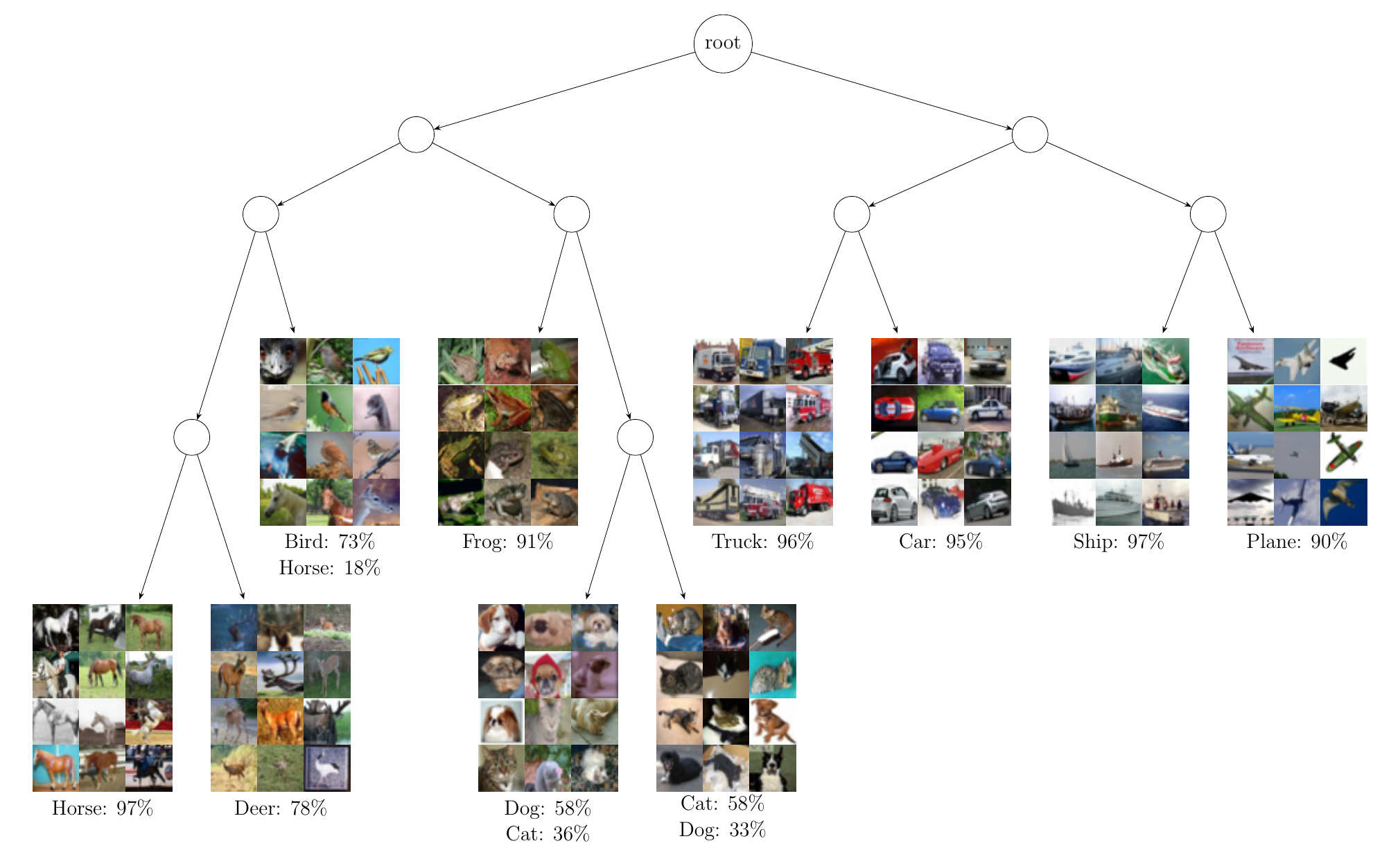}
    \caption{{\bf A tree hierarchy generated by \our{} for CIFAR-10.} There is an evident distinction into animals (left branch) and machines (right branch). Moreover, all neighbor leaves represent visually similar classes (horses and deers, dogs and cats, trucks and cars, ships and planes). Images with frogs seem to be visually similar to cats and dogs, which leads to their placement in the neighbor leaves (however cats and dogs are connected by a stronger relationship). Interestingly, a small portion of images with horses' heads are grouped together with birds because of their similar shapes. Although there is a slight mismatch between dogs and cats classes, the left leaf contains pets with bright fur photographed in front, while the right leaf includes animals with dark fur presented from the side, which coincides with our intuition. }
    \label{fig:cifar}
\end{figure}

The results presented in Table \ref{tab:large} show that \our{} outperforms the comparative methods in 3 out of 5 datasets. It gives extremely good results on CIFAR-10 and ImageNet-10, but is notably worse than CC on STL-10. We suspect that lower performance on STL-10 can be caused by inadequate choice of the backbone architecture to process images at resolution $96 \times 96$. Nevertheless, one should keep in mind that \our{} is the only hierarchical method in this comparison, and constructing a clustering hierarchy, which resembles ground truth classes, is more challenging than directly generating a flat partition.

\begin{figure}[!htb]
    \centering    \includegraphics[width=0.6\textwidth]{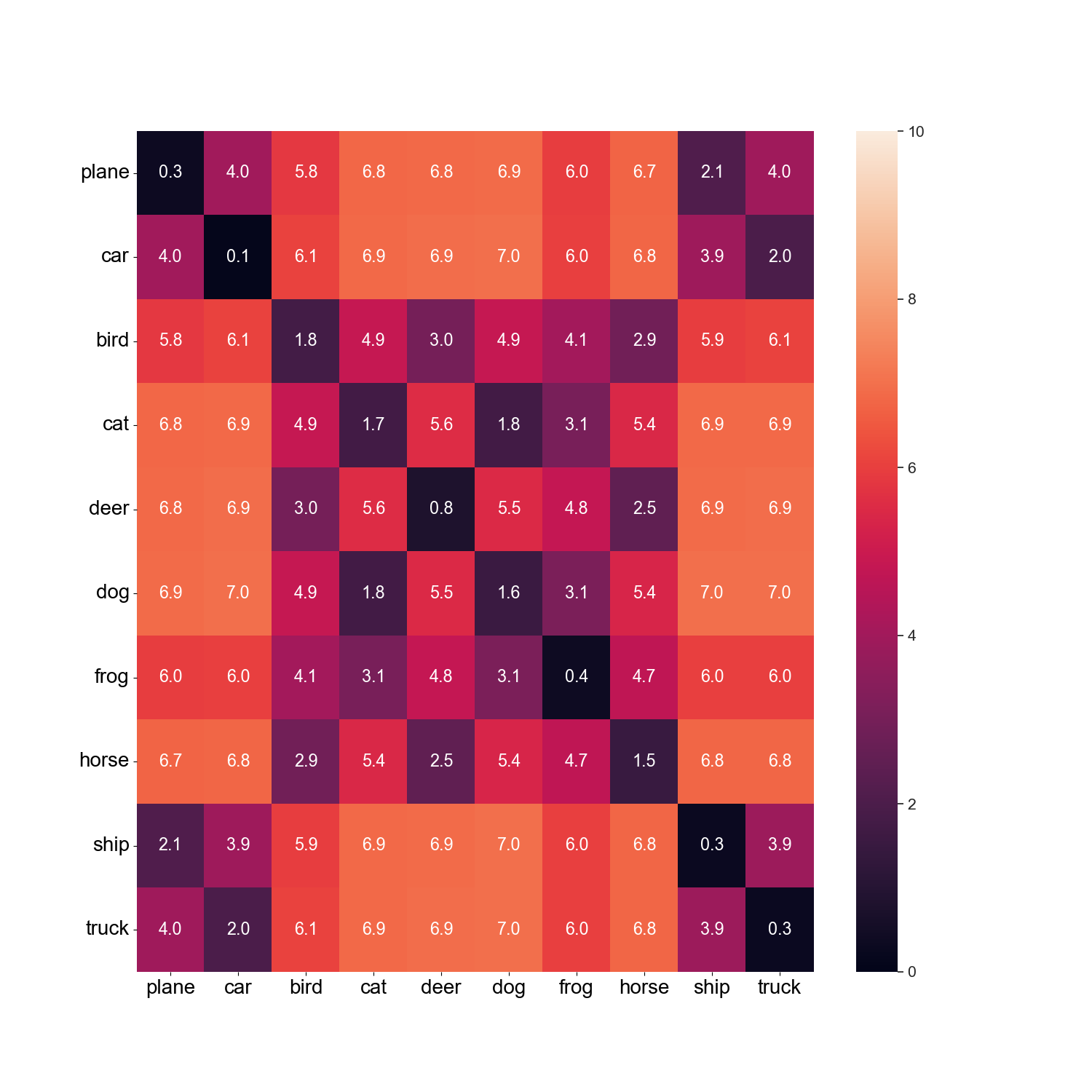}
    \caption{Distance matrices retrieved from the constructed hierarchies for ground truth classes of CIFAR-10 dataset, see text in the paper for the interpretation.}
    \label{fig:dist}
\end{figure}


\paragraph{Analyzing clustering hierarchies} To better analyze the results returned by \our{}, we plot the constructed hierarchies. Figures \ref{fig:imagenet} and \ref{fig:cifar} present and discuss the results obtained for ImageNet-10 and CIFAR-10, respectively (we refer the reader to the Appendix for more examples).  In both cases, \our{} was able to accurately model ground-truth classes as leaf nodes. In addition to the information contained in a flat partition, \our{} allows us to find relations between clusters. In particular, we observe that similar classes are localized on neighboring nodes. 

The hierarchy also allows us to define the distance $d(a,b)$ between two examples $a,b$ using the number of edges that connect them. We use the average of this score to calculate the similarity between the ground truth classes $A$ and $B$ as $d(A,B) = \frac{1}{Z} \sum_{a \in A} \sum_{b \in B} d(a,b)$, where $Z$ is the number of all pairs. The distance is small if examples from classes $A$ and $B$ are located in nearby leaf nodes (on average).

Analysis of the distance matrix in \Cref{fig:dist} confirms that objects representing the same class have the strongest relationship in the hierarchy (the diagonal entries contain the smallest values in the matrix). We also see high similarities between classes representing dogs and cats (1.8 jumps), cars and trucks (2 jumps), etc. In general, the distance matrix supports our previous findings quantitatively. 

\subsection{Analysis of the \our{} model} \label{sec:analysis}

We analyze selected properties of \our{} including the choice of the backbone network, the influence of regularization functions, the initial depth of the tree, and the training curves. Additionally, we demonstrate that training typical hierarchical methods on top of self-supervised learning models is sub-optimal. All experiments were carried out on the CIFAR-10 dataset.

\paragraph{Reliance on backbone network}

In Table \ref{tab:arch}, we show how the selection of the architecture of the base network $g$ influences the clustering results. As can be seen, \our{} gradually improves its performance with the depth of the architecture, suggesting that \our{} adapts well to deeper networks. In contrast, CC seems to be resistant to the selection of architecture and obtains optimal results on a medium-sized network.

\begin{table}[!ht]
\centering
\caption{The importance of architecture choice.
} \label{tab:arch}
\begin{tabular}{lccc|ccc}
\toprule
Method & \multicolumn{3}{c|}{\our{}}  & \multicolumn{3}{c}{CC \cite{Li2021contrastive_clustering}} \\
\midrule
Backbone & NMI & ACC & ARI & NMI & ACC & ARI\\
\midrule
ResNet18 & 0.711 & 0.768 & 0.642 & 0.650 & 0.736 & 0.569 \\
ResNet34 & 0.730 & 0.788 & 0.667 & {\bf 0.705} & {\bf 0.790} & {\bf 0.637}\\
ResNet50 & {\bf 0.767} & {\bf 0.840} & {\bf 0.720} & 0.663 & 0.747 & 0.585\\
\bottomrule
\end{tabular}
\end{table}

\paragraph{Analysis of loss function}

Next, we explain the influence of particular components of the \our{} loss function. Additionally, we verify the role of pre-training phase. As shown in Table \ref{tab:ab}, regularization functions have a significant impact on the results of \our{} boosting the NMI score by 0.2. It is also evident that pre-trainig is an essential ingredient of \our{}, which allows selecting a better starting point for building a clustering tree.

\begin{table}[!ht]
\centering
\caption{Ablation study of \our{} loss function performed on CIFAR-10. \label{tab:ab}}
\begin{tabular}{ lcccc }
\toprule
 & NMI & ACC & ARI\\
\midrule
CoHiLoss & 0.567 & 0.569 & 0.457\\
CoHiLoss + R1 & 0.629 & 0.726 & 0.549\\
CoHiLoss + R1 + R2 & {\bf 0.767} & {\bf 0.84} & {\bf 0.72}\\
\our{} w/o pre-training & 0.59 & 0.657 & 0.50 \\
\bottomrule
\end{tabular}
\end{table}

\paragraph{Selecting depth of the tree}

We investigate the choice of the initial depth of the clustering hierarchy. In \Cref{tab:simclr}, we observe a slight increase in performance by changing depth from 4 to 5. However, adding one more level does not lead to further improvement. In our opinion, using deeper trees allows for better exploration and leads to more fine-grained clusters. On the other hand, increasing the number of nodes makes optimization harder, which might explain the lower results for depth 6.

\begin{table}[!ht]
\centering
\caption{Influence of tree depth on the clustering results.} \label{tab:simclr}
\begin{tabular}{ccccc}
\toprule
 Depth & NMI & ACC & ARI\\
\midrule
4 & 0.767 & {\bf 0.840} & 0.720 &\\
5 & {\bf 0.779} & 0.839 & {\bf 0.731}\\
6 & 0.689 & 0.713 & 0.581 \\
\bottomrule
\end{tabular}
\end{table}

\paragraph{Learning curves}

\begin{figure}[!htb]
    \centering
    \includegraphics[width=0.6\textwidth]{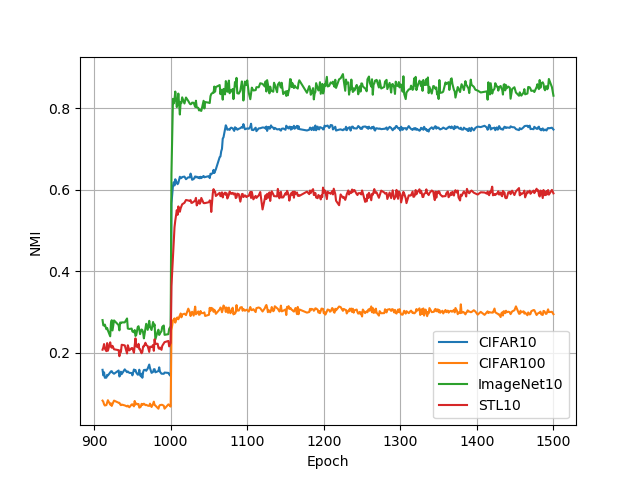}
    \caption{Learning curves on the validation set of CIFAR-10. Pre-training is performed until epoch 1000 and pruning is applied in epoch 1050. The model stabilizes its performance quickly after pruning.}
    \label{fig:wykresy}
\end{figure}

To illustrate the training phases of \our{}, we show the learning curves in \Cref{fig:wykresy}. Up to epoch 1000, we only trained the backbone model. Since in the pre-training phase, the clustering tree returns random decisions, we get a very low NMI value. After that, we start optimizing the CoHiLoss, and the NMI rapidly grows. In epoch 1050, we have a pruning phase, which results in further improvement of the NMI score. As can be seen, the model stabilizes its performance just after the pruning stage, which suggests that we can stop the training in epoch 1100. In conclusion, \our{} requires less than 100 epochs to obtain an optimal clustering score given a pre-trained model.

\paragraph{Comparison with agglomerative clustering} 

We show that our top layer responsible for constructing a decision tree is an important component of \our{} and cannot be replaced by alternative hierarchical clustering methods. For this purpose, we first train a backbone network with a typical self-supervised SimCLR technique. Next, we apply agglomerative clustering to the resulting representation. As can be seen in Table \ref{tab:agl}, agglomerative clustering gets very low results, which means that joint optimization of the backbone network and clustering tree using the proposed $\loss{}$ is a significantly better choice. In consequence, the representation taken from a typical self-supervised learning model does not provide a representation, which can be clustered accurately using simple methods.

\begin{table}[!ht]
\centering
\caption{Comparison with agglomerative clustering trained on the representation generated by the self-supervised learning model. \label{tab:agl}}
\begin{tabular}{ lcccc }
\toprule
 & NMI & ACC & ARI\\
\midrule
Agglomerative clustering & 0.265 & 0.363 & 0.147\\
\our{} & {\bf 0.767} & {\bf 0.84} & {\bf 0.72}\\
\bottomrule
\end{tabular}
\end{table}

\subsection{Comparison with hierarchical clustering methods}

To our knowledge, DeepECT \cite{mautz2019deep} is the only hierarchical clustering method based on deep neural networks. Following their experimental setup, we report the results on two popular image datasets, MNIST and F-MNIST, and consider classical hierarchical algorithms evaluated on the latent representation created by the autoencoder and IDEC \cite{guo2017improved}. In addition to the previous clustering metrics, we also use dendrogram purity (DP) \cite{kobren2017hierarchical,yang2019online}, which directly compares the constructed hierarchy with the ground-truth partition. It attains its maximum value of 1 if and only if all data points from the same class are assigned to some pure sub-tree.

\begin{table}[!htb]
    \begin{center}
    \caption{Comparison with hierarchical models in terms of DP, NMI and ACC (higher is better).
} \label{tab:hier}
    \begin{tabular}{lcccccc}
    \toprule
    Method & \multicolumn{3}{c}{MNIST} & \multicolumn{3}{c}{F-MNIST} \\
    \midrule
     & DP & NMI & ACC & DP & NMI & ACC \\
    \midrule
    DeepECT & 0.82 & 0.83 & 0.85 & 0.47 & 0.60 & 0.52 \\
    DeepECT + Aug & 0.94 & 0.93 & 0.95 & 0.44 & 0.59 & 0.50 \\
    IDEC (agglomerative complete*) & 0.40 & 0.86 & 0.85 & 0.35 & 0.58 & 0.53 \\
    AE + k-means (bisecting*) & 0.53 & 0.70 & 0.77 & 0.38 & 0.52 & 0.48 \\
    {\bf \our{}} & {\bf 0.97} & {\bf 0.97} & {\bf 0.99} & {\bf 0.52} & {\bf 0.62} & {\bf 0.65} \\
    \bottomrule
    \end{tabular}\\
    \end{center}
    \footnotesize * To report DP for flat clustering models, we use (optimally selected) typical hierarchical algorithms to build a hierarchy on the obtained data representation.
\end{table}

The results summarized in Table \ref{tab:hier} demonstrate that \our{} outperforms all baselines on both MNIST and F-MNIST datasets in terms of all metrics. Interestingly, DeepECT benefits from data augmentation in the case of MNIST, while on F-MNIST it deteriorates its performance. All methods except \our{} and DeepECT failed completely to create a hierarchy recovering true classes (see the DP measure), which confirms that there is a lack of powerful hierarchical clustering methods based on neural networks. The disproportion between the results obtained on MNIST and F-MNIST demonstrates that recovering true classes of clothes is a significantly more challenging task than recognizing handwritten digits.

\section{Conclusion}

We proposed a contrastive hierarchical clustering model \our{}, which suits well to clustering of large-scale image databases. The hierarchical structure constructed by \our{} provides significantly more information about the data than typical flat clustering models. In particular, we can inspect the similarity between selected groups by measuring their distance in the hierarchy tree and, in consequence, find super-clusters. Experimental analysis performed on typical clustering benchmarks confirms that the produced partitions are highly similar to ground-truth classes. At the same time, \our{} allows us to discover important patterns that have not been encoded in the class labels. 




%
%

\subsubsection{Acknowledgements} The research of P. Rola was supported by the National Science Centre (Poland), grant no. 2021/41/B/ST6/01370. The research of J. Tabor was supported by the National Science Centre (Poland), grant no. 2022/45/B/ST6/01117. The research of M. Śmieja was supported by the Foundation for Polish Science co-financed by the European Union under the European Regional Development Fund in the POIR.04.04.00-00-14DE/18-00 project carried out within the Team-Net program. For the purpose of Open Access, the author has applied a CC-BY public copyright licence to any Author Accepted Manuscript (AAM) version arising from this submission.

%
%
%
\bibliographystyle{splncs04}
%

\bibliography{reference}

\appendix

\section{Experimental setup}

\paragraph{Gray-scale images: MNIST and F-MNIST}

We instantiate the backbone network $g$ using ResNet18. Projection head $\pi$ used for parameterizing decisions made by internal tree nodes is a single-layer network. The projection head $\phi$ used for applying regularization $R_2$ is an identity function (NT-Xent loss is applied on the output of backbone network). We run \our{} with 16 leaves (4 tree levels). \our{} is trained for 200 epochs of pre-training and 100 epochs of tree construction (pruning is performed in epochs 10-15) using a minibatch size of $256$.


\paragraph{Color images}

For CIFAR-10, CIFAR-100 and ImageNet-10, we use ResNet50 as a backbone network. The remaining datasets of color images employed ResNet-34 network. Projection heads $\pi$ and $\phi$ are separate two-layer networks. For CIFAR-10 and CIFAR-100, we run \our{} with 32 leaves, while for the remaining datasets we use 16 leaves. \our{} is trained for 1000 epochs of pre-training and 500 epochs of tree construction using a minibatch size of $512$ for CIFAR-10, CIFAR-100 and STL-10. In case of ImageNet-10 and ImageNet-Dogs we used minibatch with size of $128$.   Pruning starts after 50 epochs of tree construction and we remove one leaf per epoch.

\paragraph{Summary of datasets} Table \ref{tab:data} summarizes datasets used for evaluation.

\begin{table}[!ht]
\centering
\caption{A summary of datasets.} \label{tab:data}
\footnotesize
\begin{tabular}{lcccc}
\toprule
 Dataset & Split & Samples & Classes & Resolution\\
\midrule
CIFAR-10 & Train+Test & 60 000 & 10 & $32 \times 32$ \\
CIFAR-100 & Train+Test & 60 000 & 20 & $32 \times 32$ \\
STL-10 & Train+Test & 13 000 & 10 & $96 \times 96$ \\
ImageNet-10 & Train & 13 000 & 10 & $224 \times 224$\\
ImageNet-Dogs & Train & 19 500 & 15 & $224 \times 224$\\
\bottomrule
\end{tabular}
\end{table}

\paragraph{Augmentations}

For all datasets, we applied four types of augmentations: random cropping, horizontal flip, color jittering, and grayscaling.

\section{Dendrogram purity for datasets of color images}

In Table \ref{tab:den}, we present dendrogram purity obtained by \our{} for datasets of color images. This benchmark can be used to compare other hierarchical clustering models.

\begin{table}[!ht]
\centering
\caption{Dendrogram purity on color images.} \label{tab:den}
\begin{tabular}{cc}
\hline
 & DP  \\
\hline
CIFAR-10 & 0.715\\
CIFAR-100 & 0.231 \\
STL-10 & 0.530 \\
ImageNet-10 & 0.895  \\
ImageNet-Dogs & 0.190 \\
\hline
\end{tabular}

\end{table}

\section{Additional results}

We present hierarchies and distance matrices constructed for the remaining datasets in the following figures.

\begin{figure}[!ht]
    \centering
    \includegraphics[width=\textwidth]{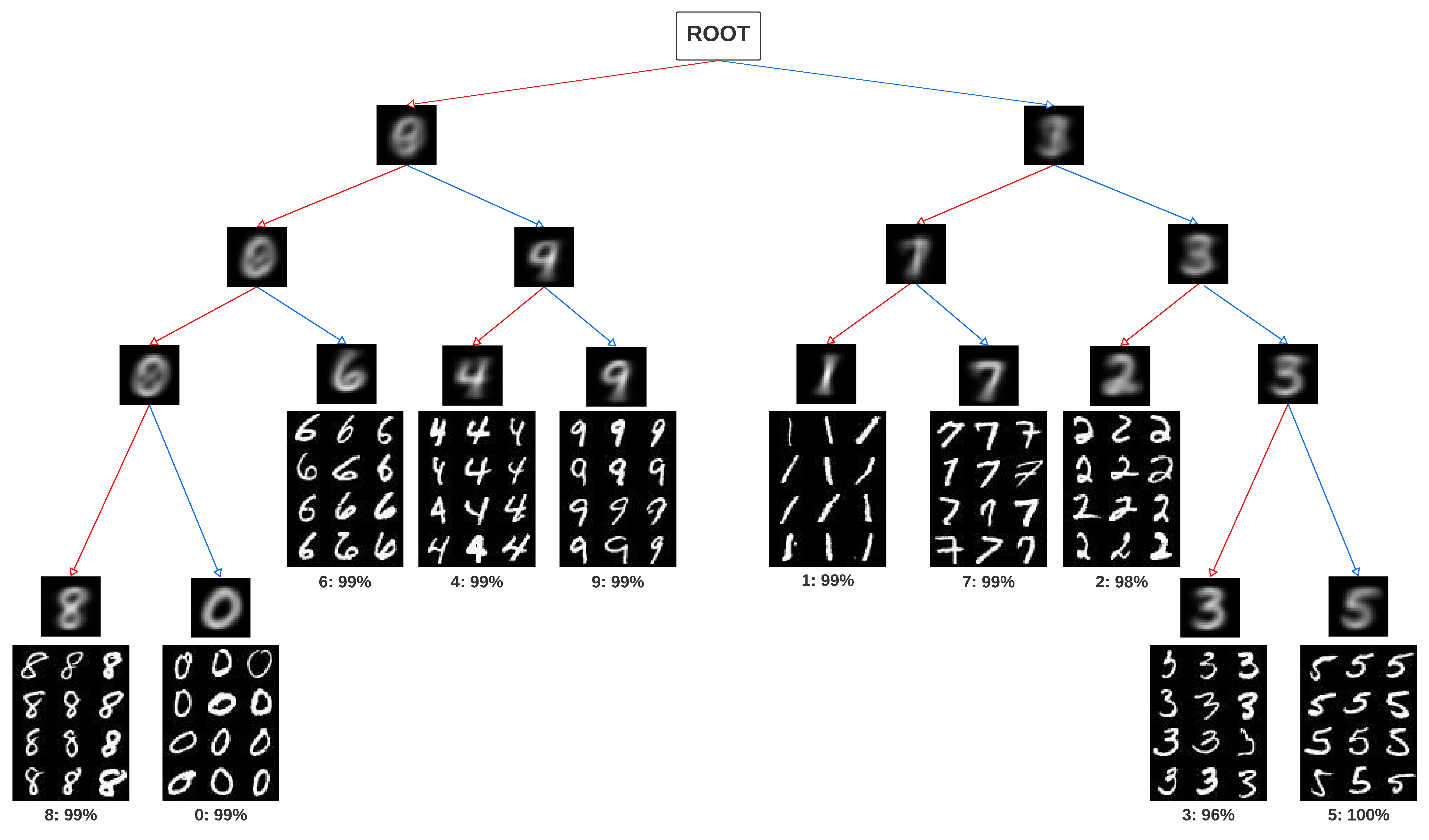}
    \caption{ {\bf Tree hierarchy constructed for MNIST.} Observe that neighboring leaves contain images of visually similar classes, e.g. 8 and 0; 4 and 9; 1 and 7. Such a property holds also for nodes in the upper levels of the tree -- the left sub-tree contain digits with circular shapes, while the digits located in the right sub-tree consist of lines.}
    \label{fig:mnist}
\end{figure}

\begin{figure}[!htb]
    \centering
    \includegraphics[width=0.85\textwidth]{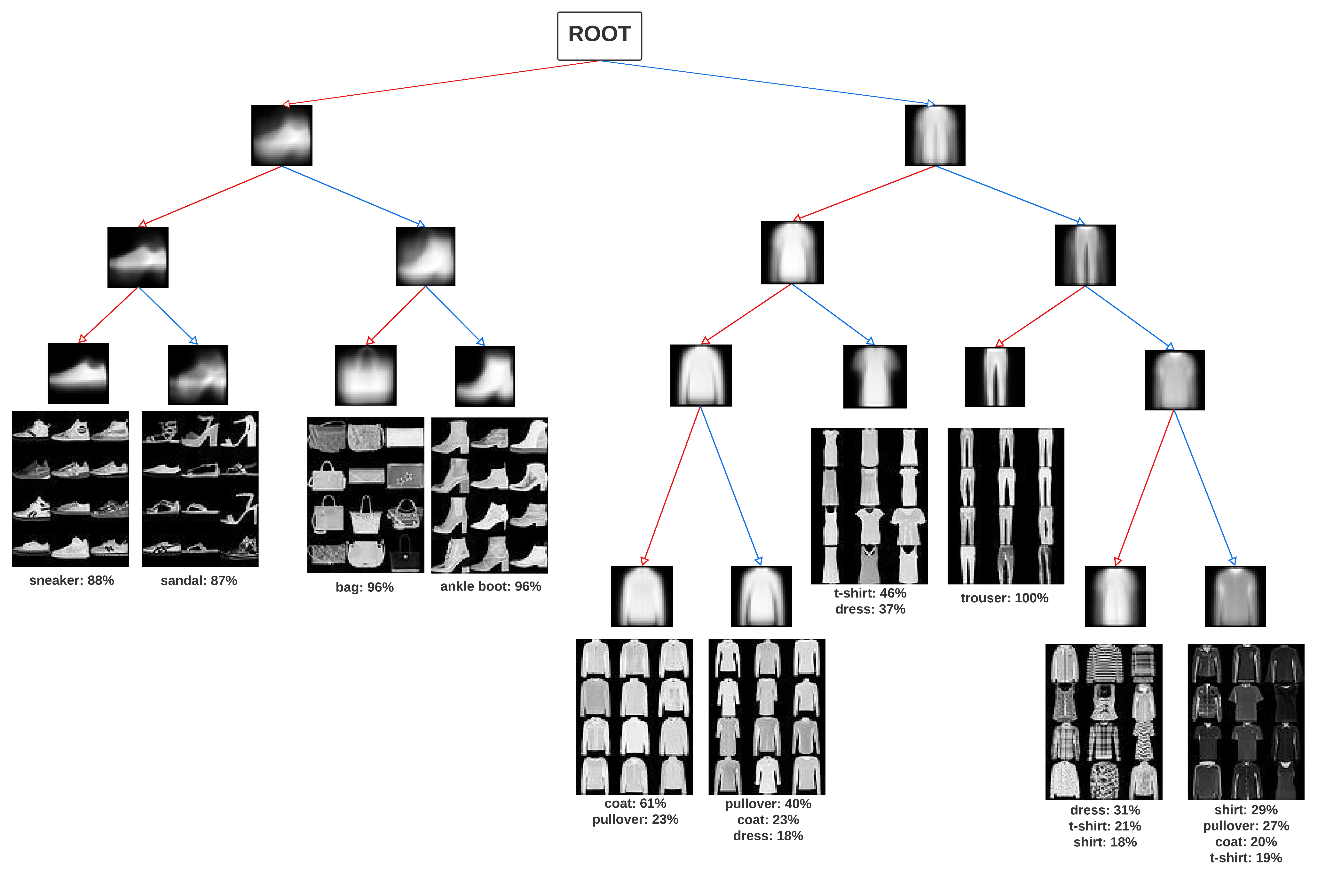}
    \caption{{\bf Tree hierarchy generated by \our{} for F-MNIST} (images in the nodes denote mean images in each sub-tree). The right sub-tree contains clothes while the other items (shoes and bags) are placed in the left branch. Looking at the lowest hierarchy level, we have clothes with long sleeves grouped in the neighboring leaves. The same holds for clothes with designs. Observe that \our{} assigned white-colored t-shirts and dresses to the same cluster, while trousers are in the separate one. Small shoes such as sneakers or sandals are considered similar (neighboring leaves) and distinct from ankle shoes. Concluding, \our{} is able to retrieve meaningful information about image semantics, which is complementary to the ground truth classification.}
    \label{fig:fmnist}
\end{figure}

\begin{figure}
    \centering
    \includegraphics[width=\textwidth]{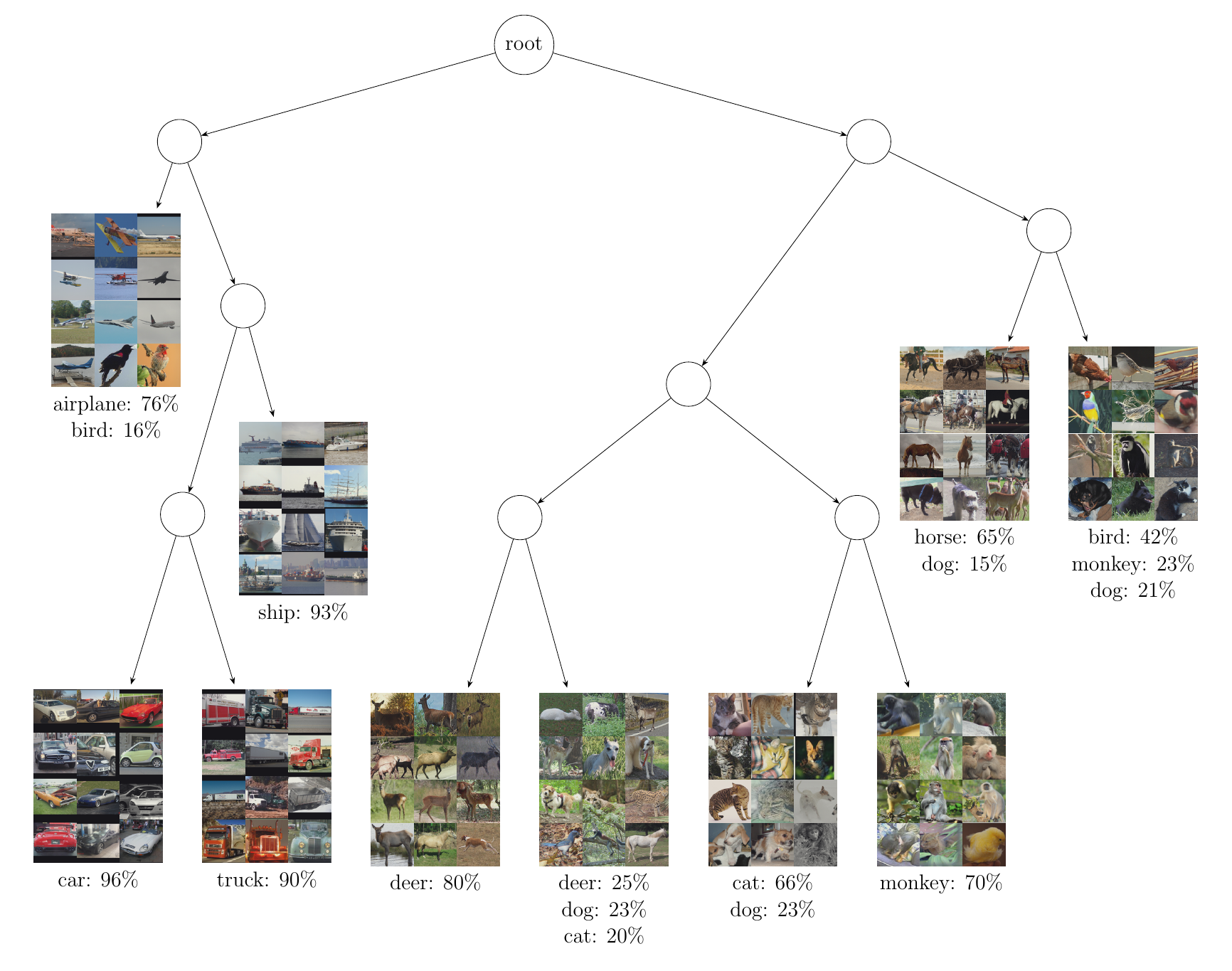}
    \caption{{\bf Tree hierarchy generated by \our{} for STL-10}}
    \label{fig:stl10}
\end{figure}

\begin{figure}[!ht]
    \centering
    \subfloat[MNIST]{\includegraphics[height=5.2cm]{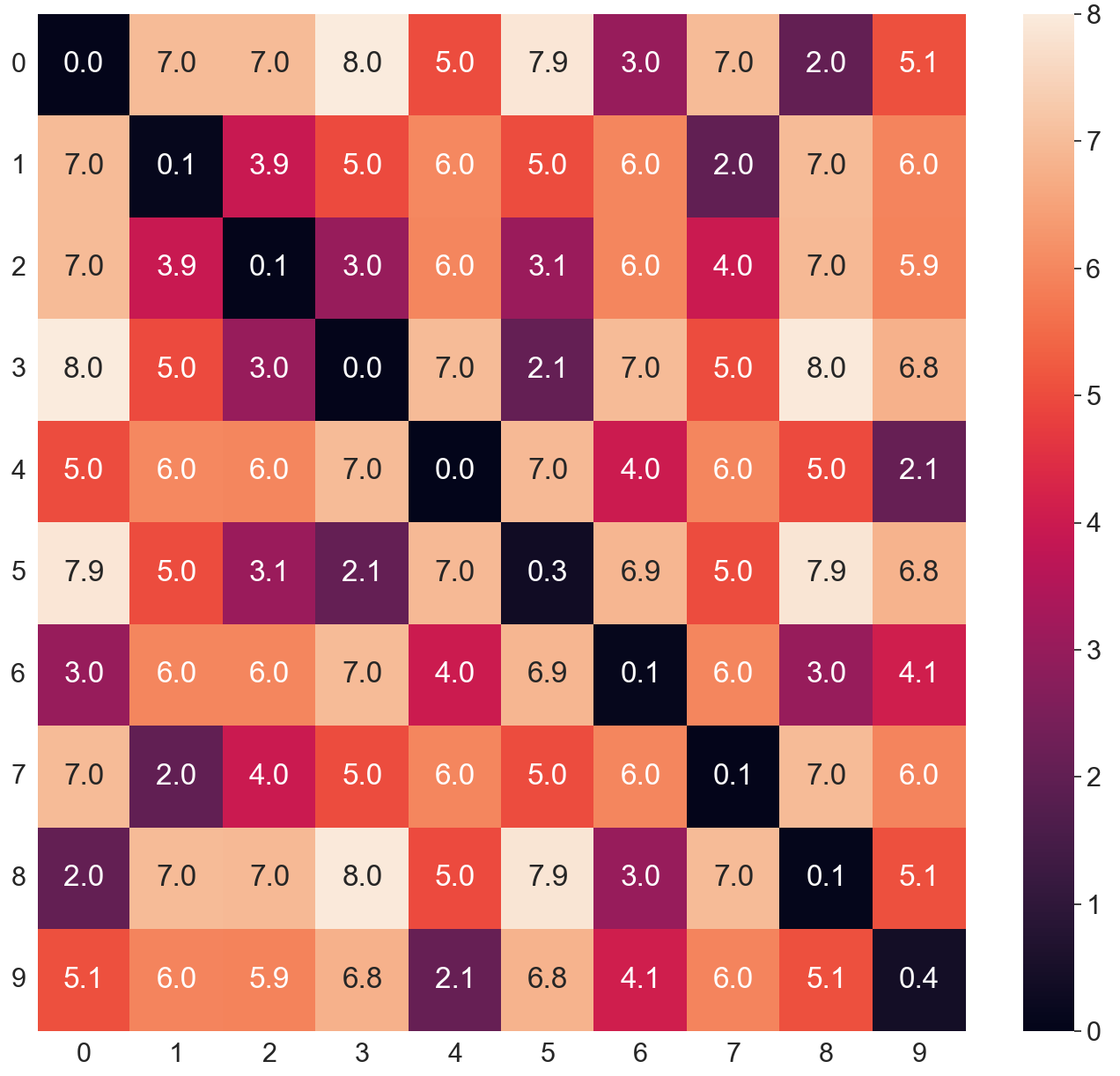}
    }
    \subfloat[F-MNIST]{\includegraphics[height=5.2cm]{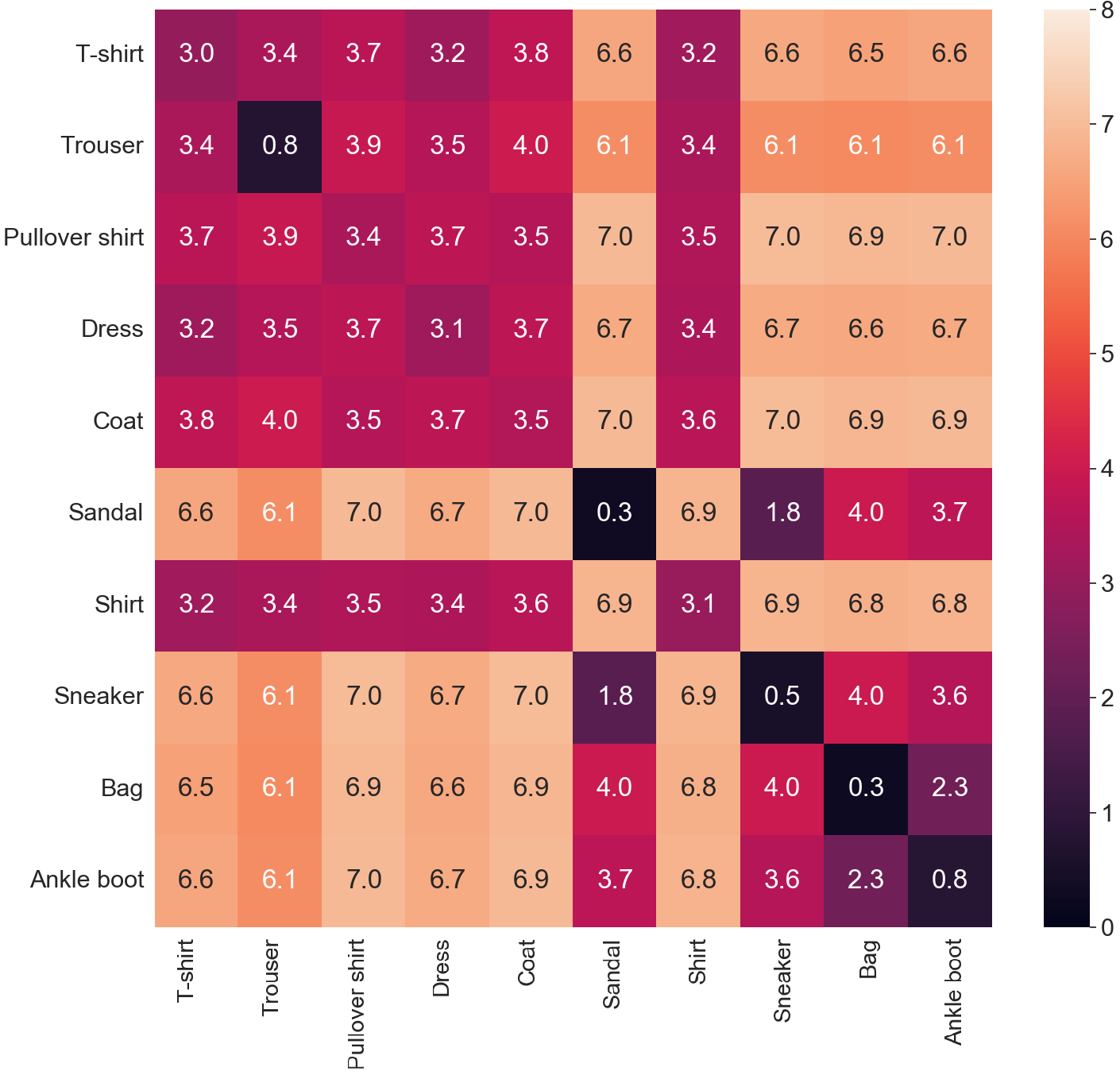}
    }\\
    \subfloat[STL-10]{\includegraphics[height=5.2cm]{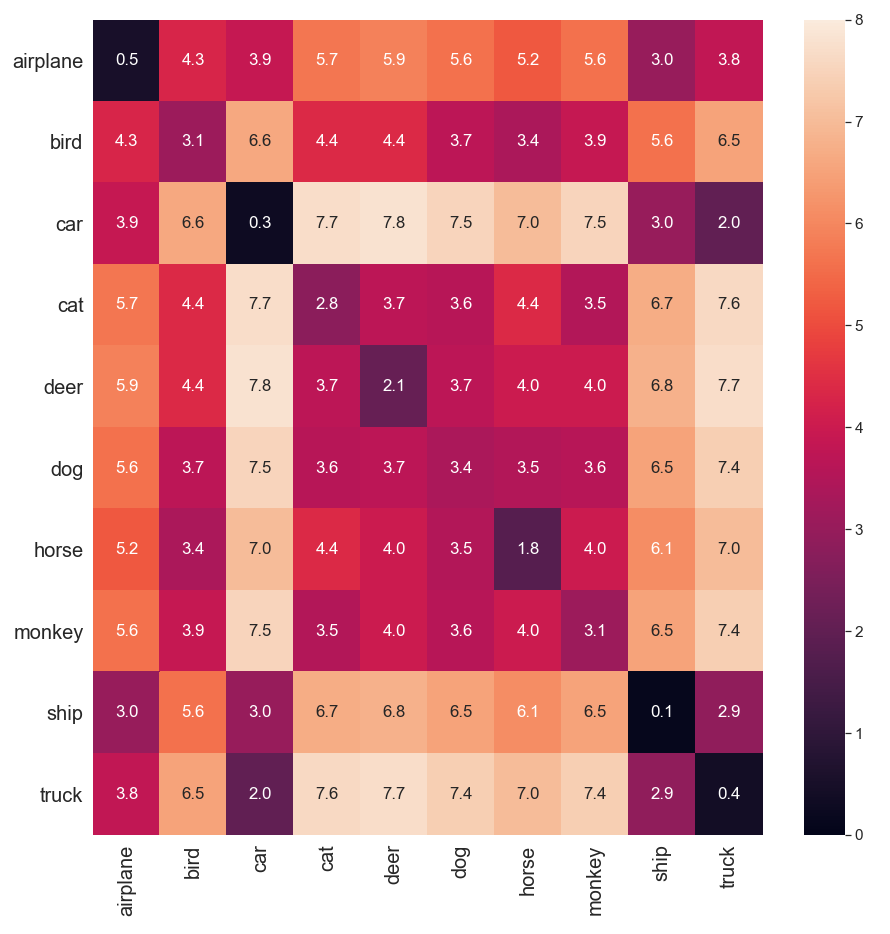}
    }
    \subfloat[ImageNet-10]{\includegraphics[height=5.2cm]{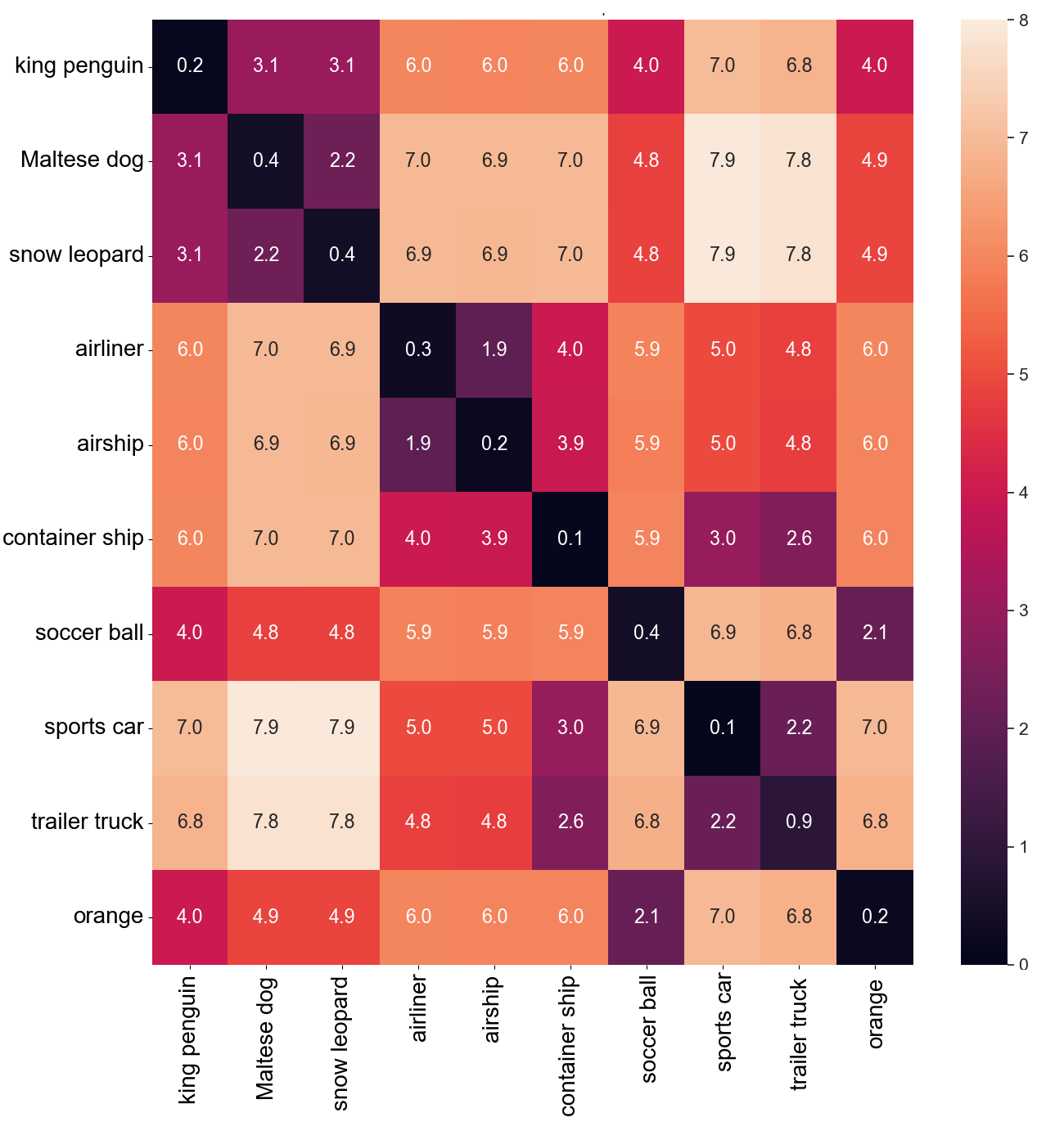}
    }\\
    \subfloat[ImageNet-Dogs]{\includegraphics[height=5.2cm]{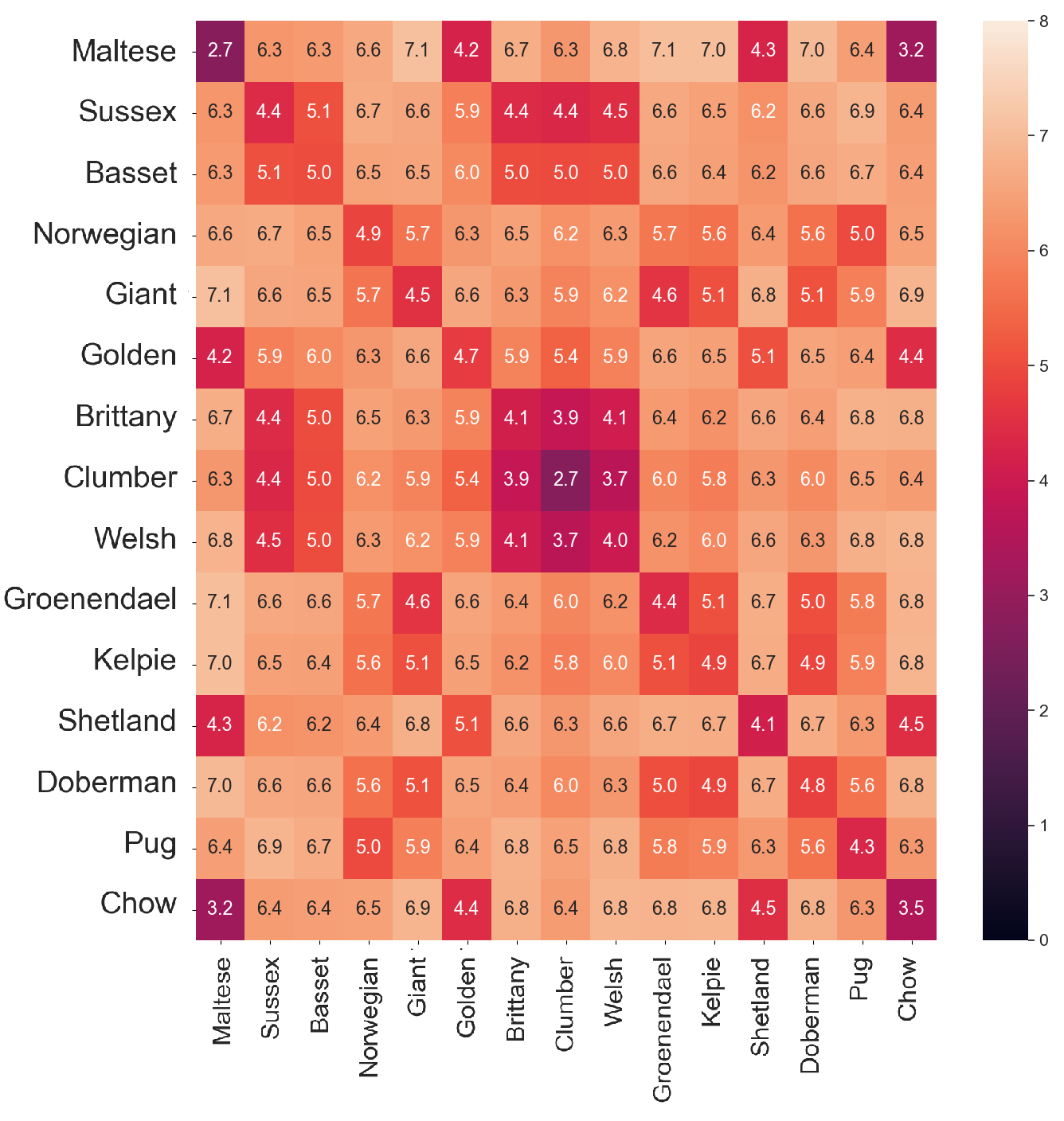}
    }
    \subfloat[CIFAR-100]{\includegraphics[height=5.2cm]{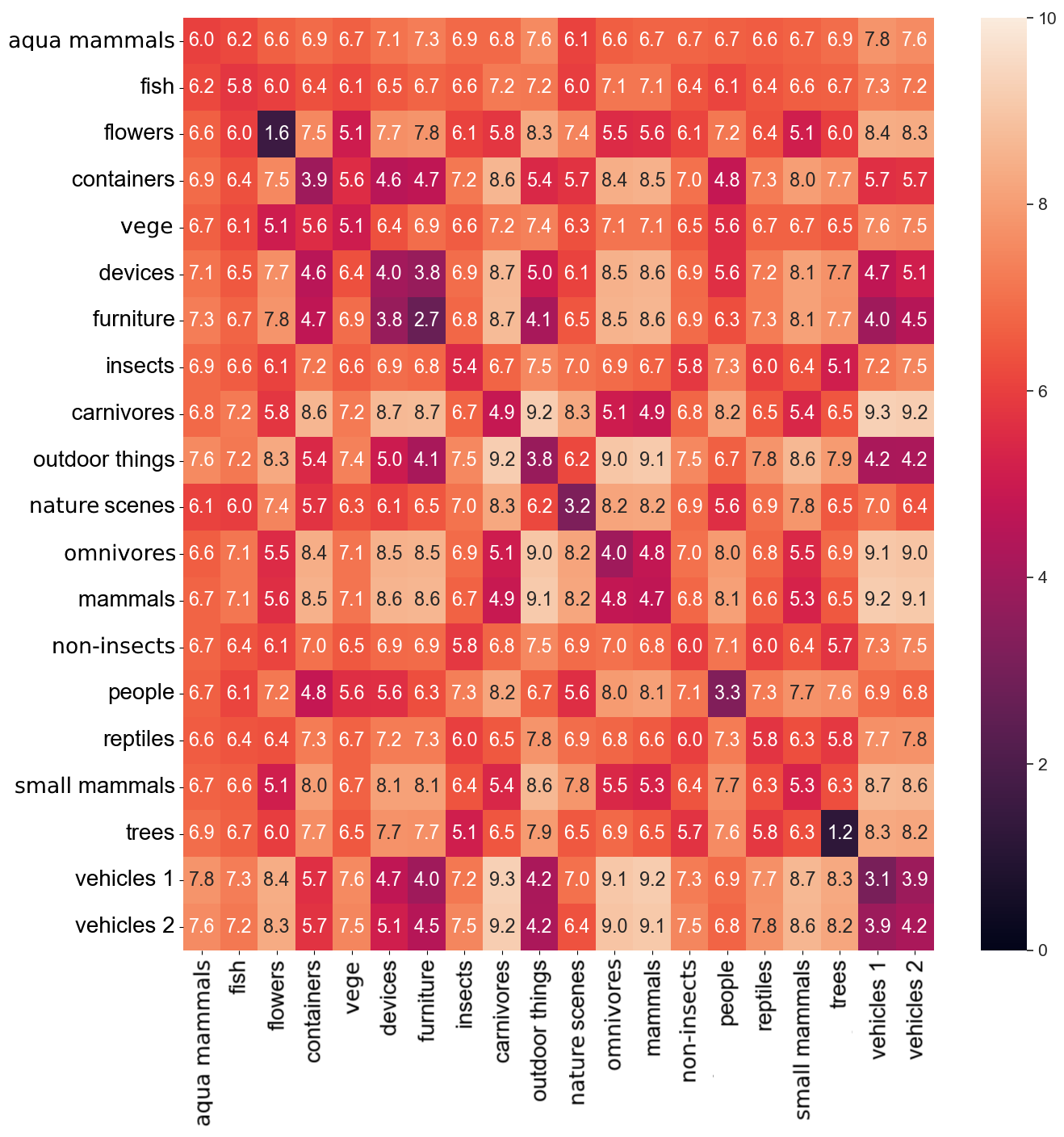}
    }
    \caption{Distance matrices retrieved from the constructed hierarchies for ground truth classes.}
    \label{fig:dist2}
\end{figure}

\end{document}